\documentclass[letterpaper, 10 pt, conference]{ieeeconf}  

\IEEEoverridecommandlockouts                              

\usepackage{threeparttable}
\usepackage[T1]{fontenc}
\usepackage{amsmath}
\usepackage{hyperref}
\usepackage{siunitx}
\usepackage{booktabs}
\usepackage{amssymb}
\usepackage{graphicx}
\usepackage{rotating}
\usepackage{caption}
\usepackage{multirow}
\usepackage{bbding}
\usepackage{pifont}
\usepackage{cite}
\usepackage{amsmath,amssymb,amsfonts}
\usepackage{algorithmic}
\usepackage{textcomp}
\usepackage{xcolor}
\usepackage{authblk}
\title{\LARGE \bf
LitSim: A Conflict-aware Policy for Long-term Interactive Traffic Simulation
}

\author[{$\dagger$}]{Haojie Xin \thanks{Haojie Xin is with the Department of Computer Science and Technology, Xi'an Jiaotong University, {\tt\small pinkman@stu.xjtu.edu.cn}}}
\author[{$\ddagger$}{*}]{Xiaodong Zhang}
\author[{$\S$}]{Renzhi Tang}
\author[{$\dagger$}]{\\Songyang Yan}
\author[{$\dagger$}]{Qianrui Zhao}
\author[{$\dagger$}]{Chunze Yang}
\author[{$\ddagger$}{$\S$}]{Wen Cui}
\author[{$\dagger$}{$\S$}{*}]{Zijiang Yang \thanks{{*}Corresponding author: {\tt\small Zhangxiaodong@xidian.edu.cn, zijiang@xjtu.edu.cn}}}
\affil[{$\dagger$}]{Xi'an Jiaotong University, Xi'an, China}
\affil[{$\ddagger$}]{Xidian University, Xi'an, China}
\affil[{$\S$}]{Synkrotron, Inc., Xi’an, China}

\begin{document}
\maketitle
\thispagestyle{empty}
\pagestyle{empty}

\begin{abstract}
Simulation is pivotal in evaluating the performance of autonomous driving systems due to the advantages of high efficiency and low cost compared to on-road testing. Bridging the gap between simulation and the real world requires realistic agent behaviors. However, the existing works have the following shortcomings in achieving this goal:~(1) log replay offers realistic scenarios but often leads to collisions due to the absence of dynamic interactions, and~(2) both heuristic-based and data-based solutions, which are parameterized and trained on real-world datasets, encourage interactions but often deviate from real-world data over long horizons. In this work, we propose LitSim, a long-term interactive simulation approach that maximizes realism by minimizing the interventions in the log. Specifically, our approach primarily uses log replay to ensure realism and intervenes only when necessary to prevent potential conflicts. We then encourage interactions among the agents and resolve the conflicts, thereby reducing the risk of unrealistic behaviors. We train and validate our model on the real-world dataset NGSIM, and the experimental results demonstrate that LitSim outperforms the currently popular approaches in terms of realism and reactivity.

\end{abstract}

\section{INTRODUCTION}
The rapid development of autonomous driving technology is exciting as it promises tremendous social and economic benefits by reducing traffic accidents and improving traffic efficiency. According to an estimate from~\cite{kalra2016driving}, proving that Automated Driving Systems (ADSs) can reduce traffic accidents by 20\% compared to human drivers would require approximately 8 billion kilometers of on-road testing. In practice, simulation testing can significantly reduce the cost of testing an ADS from the on-road testing in cities~\cite{importancesimulation}~\cite{tang2023survey}. The key problem of achieving this is how to simulate traffic scenarios like real-world traffic which can have reasonable interactions over long horizons. However, it is challenging to generate realistic and long-term traffic scenarios due to \textit{simulation drift} and \textit{fragmented scenarios}.
\begin{figure}[t]
\centering
\includegraphics[width=\linewidth]{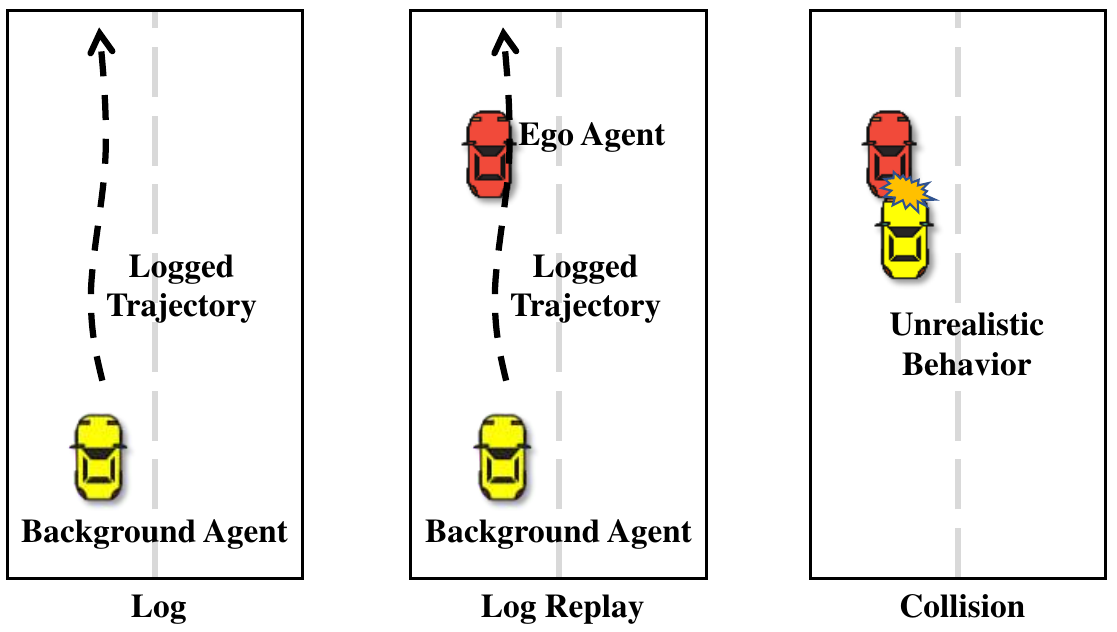}
\caption{{{\bf An example of simulation drift. }The logged trajectory determines the motion of the background agent, which fails to react to the entry of the ego agent, thereby resulting in a collision and an unrealistic scenario.}}
\label{unrealistic_example}
\end{figure}

Currently, many traditional simulators are based on log replay, e.g., SUMO~\cite{lopez2018microscopic}, Carla~\cite{dosovitskiy2017carla}. Although the movement of background agents maintains fidelity to the real world, this approach often overlooks the interactive behaviors and kinematic constraints between the ego agent and other background agents. Log replay will lead to collisions, and such unrealistic behaviors are a result of what is called simulation drift, as shown in Fig.~\ref{unrealistic_example}. A common method to address this issue involves heuristic-based models, such as the Intelligent Driver Model~(IDM)~\cite{treiber2000congested}~\cite{kesting2007general}. However, the generated scenarios lack complex interaction behavior between the ego agent and background agents due to handcrafted rules. Consequently, background agents exhibit simplistic and repetitive behaviors, such as car-following, over extended periods.

To address the shortcomings of traditional simulators, some data-driven methods have tried to create realistic and diverse scenarios from real-world data. They allow for the interactions between the ego agent and background agents, demonstrating significant potential for closed-loop evaluation. For example, SimNet~\cite{bergamini2021simnet} formulates the simulation problem as a Markov process and leverages deep neural networks to predict future states. TrafficSim~\cite{suo2021trafficsim} parameterizes a policy with an implicit latent variable that generates plans for all agents. However, these approaches model the simulation as a motion prediction problem, which will lead to deviations from real-world data over long-term horizons due to compounding errors. Some approaches~\cite{ho2016generative}~\cite{rail}~\cite{2022gail} based on Generative Adversarial Imitation Learning~(GAIL) have been proposed to learn driving strategies directly from real-world data. GAIL-based approaches suffer from mode collapse, where the generator fails to capture the full diversity of expert behavior. Consequently, this can lead to unrealistic behaviors, such as collisions or driving off-road. Additionally, the collected data of the above methods is from open driving datasets, such as the Waymo Open Dataset~\cite{sun2020scalability} and Argoverse~\cite{chang2019argoverse}, which are often fragmented. For instance, in the Waymo Open Dataset, only 30\% of the scenarios last more than 10 seconds, making it challenging to simulate the long-term traffic flow.
    
In this work, we propose an approach called LitSim to generate interactive and realistic scenarios in long-term simulations. LitSim identifies potential conflicts in traffic simulations and then employs a conflict-aware policy to take over background agents. Thus, LitSim only intervenes once a potential conflict is detected. It generates interactive behavior during the takeover to ensure that the scenarios do not deviate from real-world data. When conflicts are avoided, LitSim switches back to the replay policy. Compared to existing approaches, LitSim excels in terms of realism and reactivity and demonstrates the advantages of long-term simulations. Specifically, our contributions are as follows:
    
\begin{itemize}
\item We propose a learning-based long-term interactive simulation method, called LitSim, to maintain maximum realism by minimizing our interventions in the raw log, only stepping in when necessary, such as to prevent potential conflicts.

\item We identify potential conflicts by jointly predicting the future motions of agents, considering their interactions. We also leverage a conflict-aware policy to roll out consistent plans across various interaction scenarios.

\item We have implemented LitSim based on SUMO, a widely used simulator. We train and evaluate our method on the NGSIM dataset. The experimental results show that LitSim outperforms three popular approaches (i.e., IDM, GAIL, and SimNet).
\end{itemize}

\section{Related Work}

\subsection{Motion prediction}
Motion prediction involves forecasting an agent's future motion based on its past states and contextual information. Traditional approaches track an object and propagate its state to predict its future motion (e.g., Particle Filter~\cite{gustafsson2002particle} with Dynamic Bicycle Model~\cite{dynamicvehicle}). Recently, machine learning-based methods have gained popularity due to their ability to account for physical factors, road conditions, and interactions, adapting to complex scenarios. These methods are achieved through various representation methods, such as vector representations~\cite{gao2020vectornet} and rasterized images~\cite{gilles2021home}. Due to the inherent uncertainty in human intent, future distribution exhibits multimodality. To explain it, some researchers use generative models~\cite{generative1}~\cite{generative2}, goal targets~\cite{sun2022m2i}~\cite{qcnet}, or maneuvers~\cite{maneuvre}~\cite{zhang2023intention} to generate low-level trajectories that are conditioned on the intent.

Building on recent advancements, LitSim includes a neural network that enhances motion prediction for simulated agents. This network processes diverse inputs to improve prediction accuracy and effectively manages complex interactions in dynamic environments.

\subsection{Intelliegnt Agent Model}
An authentic human driver model is vital for accurate driving simulations, offering substantial benefits for automotive safety research. Traditionally, human driver modeling relies on heuristic-based approaches, such as the Intelligent Driver Model~\cite{treiber2000congested}\cite{kesting2007general}, which generate the driving behaviors but are limited to handcraft. Recent methods attempt to mimic expert behavior through direct strategy optimization. Imitation Learning~(IL) aims to learn control strategies from experts. Behavioral cloning\cite{bergamini2021simnet} involves training a model to map inputs to corresponding expert's state-action pairs, enabling the learner to replicate the demonstrated behavior. However, this approach can result in causal confusion~\cite{casual} and distribution shift~\cite{limitationofBC}. GAIL-based approaches~\cite{rail}\cite{2022gail}\cite{igl2022symphony} are robust to demonstration perturbations. These approaches filter out noise in the expert data, leading to accurate policy learning. Several methods~\cite{rl1}\cite{rl2} have been proposed to apply Reinforcement Learning~(RL) to autonomous driving. However, it is challenging to handcraft a precise and accurate reward function that trains the agent behavior like any particular human driver.

In this work, We combine IL and RL to train a behavior policy from real-world data. It generalizes to different traffic scenarios and reproduces the human drivers' behaviors, such as performing emergency braking.
\begin{figure*}[t]
\centering
\includegraphics[width=\textwidth]{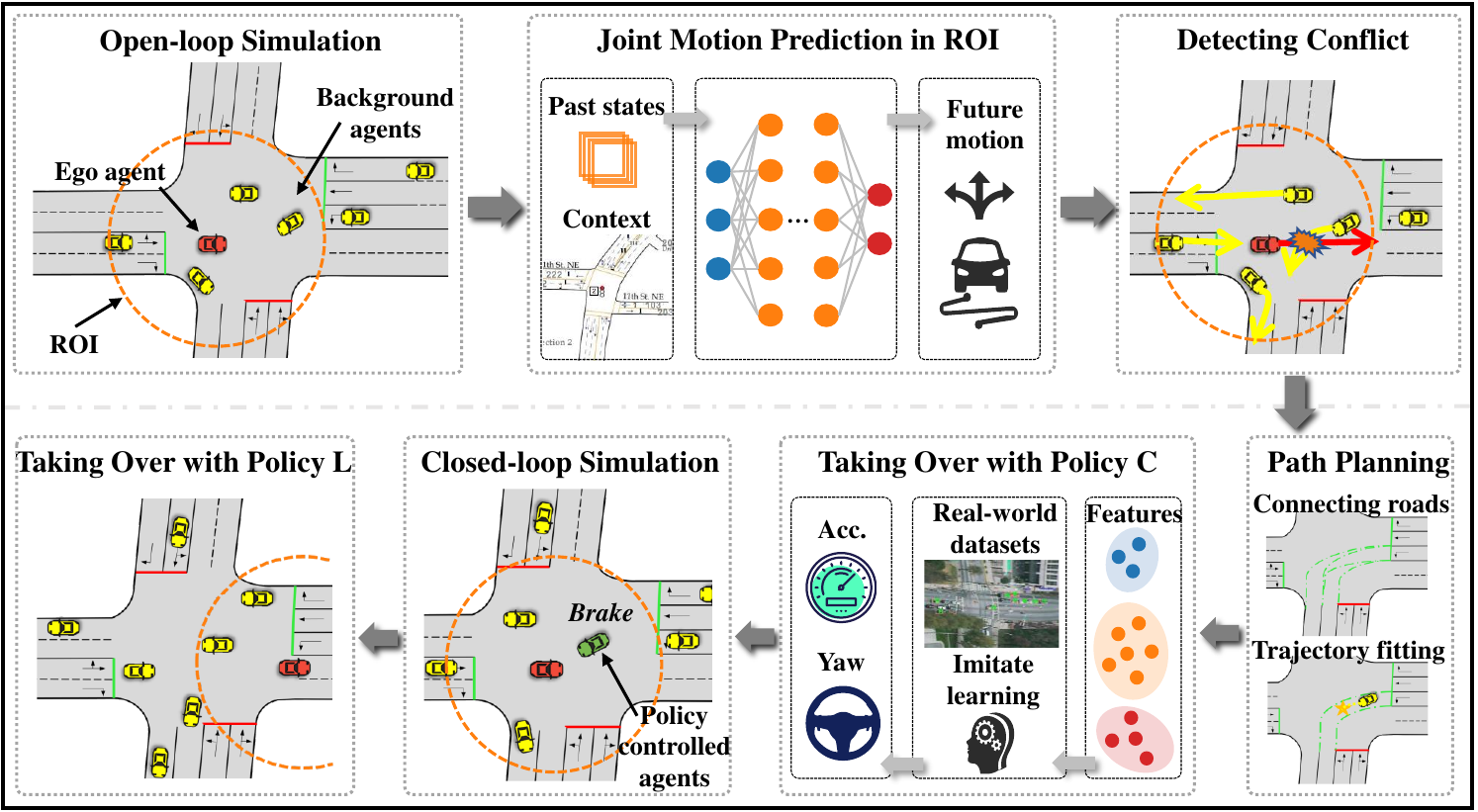}
\caption{{{\bf The Overview of LitSim.} Beginning with a basic log replay in an open-loop simulation, LitSim employs a neural network to predict the future motions of agents within the ROI. These motions will be utilized to detect potential conflicts. Once a potential conflict is detected, we formulate new trajectories for any involved background agents. Then, we apply the conflict-aware policy $C$ to guide background agents along the new path. For instance, a brake initiation in this example helps avoid a collision with the ego agent, thereby ensuring the closed-loop simulation runs smoothly without any interruptions. After resolving all conflicts, background agents switch back to the log-replay-based policy, i.e., policy $L$. }}
\label{pipeline}
\end{figure*}

\subsection{Traffic Simulation}
Traffic simulation needs to provide realistic and interactive simulations for the evaluation of an ADS. Recently, some learning-based approaches have trained from real-world data, serving as data-driven simulators with significant potential for closed-loop evaluation. For instance,
SimNet~\cite{bergamini2021simnet} trains neural networks using driving data to generate realistic scenarios. It primarily focuses on generating future trajectories for individual agents, ignoring potential collisions in traffic-intensive scenarios. To address this, SafetyNet~\cite{vitelli2022safetynet} introduces feasibility checks for generated trajectories to filter out collision-prone trajectories. However, this may inadvertently exclude collisions that are unavoidable in the real world. InterSim~\cite{sun2022intersim} predicts inter-agent relationships explicitly in traffic scenarios, thereby generating consistent trajectories for multiple agents. MixSim~\cite{suo2023mixsim} models the goals of agents as routes within the road network, training a reactive, route-conditional policy that explicitly incorporates these goals.

Compared to existing models, we propose a long-term interactive simulation approach that maximizes realism and encourages interactions by minimizing our interventions in the raw log.

\section{problem formualtion}
Our method aims to simulate the multiple agents within a given High-Definition~(HD) map denoted as $M$, and take into account dynamic states of $X$ for $N$ traffic agents. Our goal is to simulate their future motion $Y$ forward up to a finite horizon $T$. When background agents perform a simulation based on log-replay, it is essential to determine in real time whether conflicts with the ego agent will arise in the future. Once conflicts are detected, a behavior policy takes over the background agents in advance to avoid collisions. Otherwise, the simulation is still performed according to log-replay. After resolving all detected conflicts, the background agents revert to the log-replay-based policy. Thus, the problem can be formally represented as:
\begin{equation}
\begin{aligned}
Y = \begin{cases}
C(X,M), & \textit{if $D$ = True \& $t\leq s$} \\
L(X,M), & \textit{if $D$ = True \& $s< t \leq T$} \\
L(X,M), & \textit{if $D$ = False}
\label{eq1}
\end{cases}
\end{aligned}
\end{equation}
where $C$, $L$, and $D$ stand for the conflict-aware policy, log-replay policy, and whether there are potential conflicts between agents in the future.

In this work, We use $ X_t = \{x_t^1, x_t^2,...,x_t^N\} $ to represent the joint dynamic states of N agents at time $t$, where $x_t^n$ is the state of the \textit{n}-th agent at time $t$. Specifically, each agent state is parameterized $x_t^n = \{d_x,d_y,d_w,d_h,d_\theta,d_v\}$ with 2D coordinates, width, height, yaw and velocity. With a slight abuse of notation, we use superscripted $ X^n = \{x_{t-\tau}^n,x_{t-\tau+1}^n,...,x_t^n\}$ to denote the past states of the \textit{n}-th agent. $ X = X_{t-\tau:t}^{1:\textit{N}} $ denote the joint agents states from time $t-\tau$ to $t$, where $\tau$ is the observed past steps. Similar to $X$, $Y = Y_{t+1:t+T}^{1:N}$ denote the joint future states from time $t+1$ to $t+T$, where $T$ is a finite future horizon. We use $D = D_{t+s}^{1:N}$ to denote whether there are potential conflicts at time $t+s$, where $s$ is the step in which the conflict occurs in the future. To avoid conflicts in future simulations, we use control policies to simulate background agents as shown in~(\ref{eq1}). Therefore, the future states of these background agents can be represented as $ Y_{t+1:t+s}^{1:j}=C(X,M)$. Then, we use $ Y_{t+s+1:t+T}^{1:j}=L(X,M)$ to represent their future states. For agents without conflicts, the future states are represented as $ Y_{t+1:t+T}^{j+1:N}=L(X,M) $. In the following section, we control background agents by alternating between policies $C$ and $L$ that can realistically forward such reactive episodes. 

The core problems to be solved in this work are 1) to design an accurate motion prediction algorithm to detect collision $D$, 2) to generate a conflict-aware policy $C$ before the collision occurs, and 3) thereby to build a complete interactive simulation $Y$.

\section{method}
The overview of our simulation method is shown in Fig.~\ref{pipeline}. In this section, we will describe joint motion prediction with interaction, conflict detection, and conflict-aware control policy in detail. The overall simulation section explains how to iterate the above processes in each simulation step.
\begin{figure}[t]
\centering
\includegraphics[width=\linewidth]{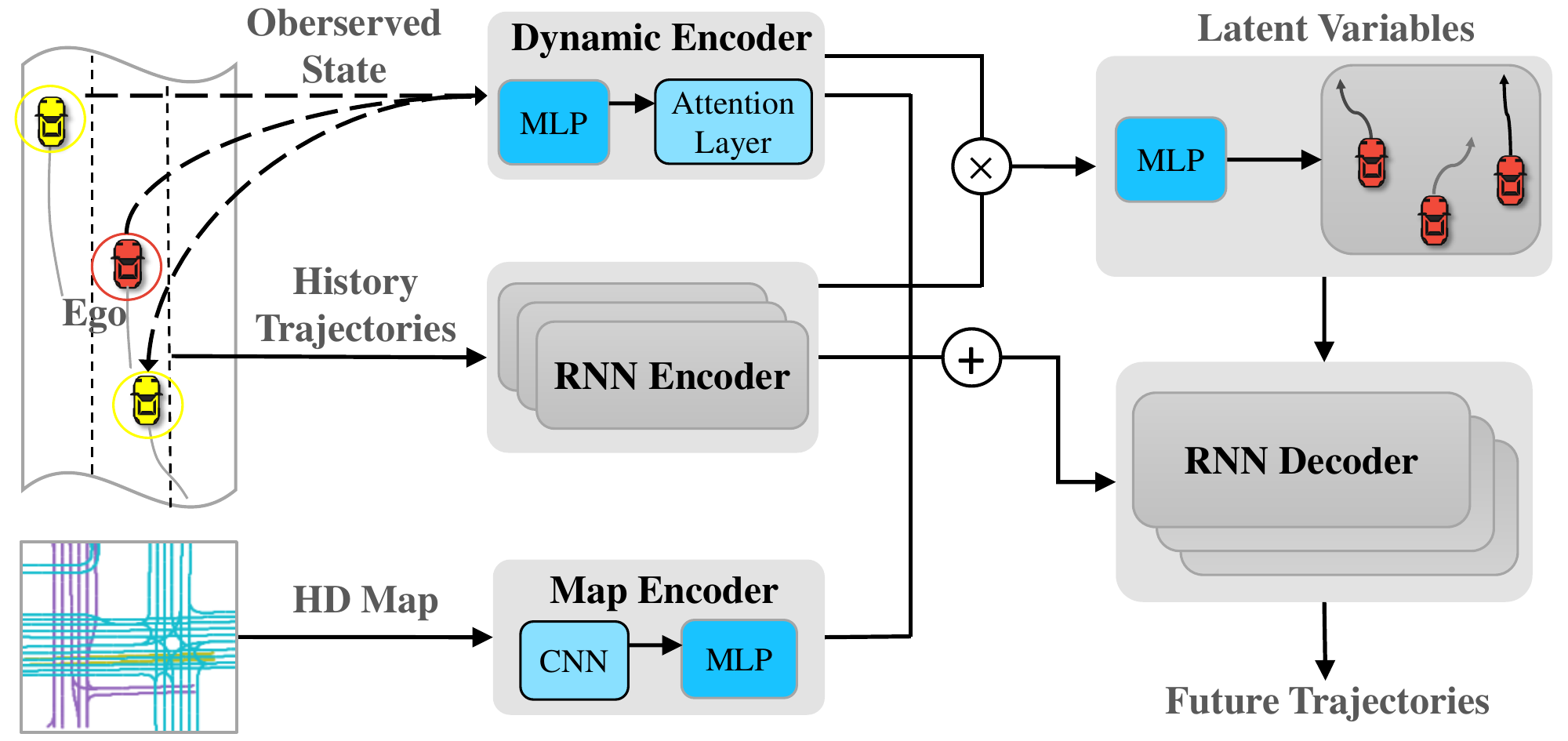}
\caption{{The architecture of predictor}}
\label{predictor}
\end{figure}
\subsection{Joint Motion Prediction with Interaction} 
\label{Joint Motion Prediction with Interaction}
LitSim predicts not only the future states for background agents but also those of the ego agent. The reason is that obtaining the ego agent's plan from a specific future horizon is challenging without a god's-eye view perspective at the simulator level. 

We tackle motion prediction through a probabilistic approach for continuous space but discrete time system with a finite (yet variable) number of $N$ interacting agents. Our predictor is based on a deep neural network with three types of encoders and a decoder. The architecture is shown in Fig.~\ref{predictor}. The inputs of the three encoders are an HD map, history trajectories, and current observed states at time $t$, and their outputs are fed to the decoder. The first encoder extracts the map features by encoding the rasterized images of the HD map with Convolutional Neural Networks~(CNNs). The CNNs are followed by a Multi-layer Perceptron~(MLP) to learn important information from map features. The second encoder embeds 2D coordinates of the history trajectories into a representation of high-dimensional vectors based on Recurrent Neural Networks~(RNNs). The third encoder, composed of an MLP, encodes the current observed states and is also known as the dynamic decoder. It utilizes multi-head attention mechanisms~\cite{vaswani2017attention} to encode interactions between agents. The dynamic encoder only calculates the current observed states at time $t$ rather than those before time $t$, as the computational cost is high. The observed states are projected into a high-dimensional space to represent the interactions, and the interaction vectors are concatenated with the historical vector and map features. We introduce a set of latent variables to learn agent maneuvers~(e.g., keep lane or lane change) from historical and interaction vectors. Finally, the concatenated tensors are fed to the decoder RNNs to predict the future trajectories conditioned on the learned maneuvers. We train the predictor by minimizing the Negative Log-Likelihood loss~(i.e., the difference between the predicted probability distribution and the ground truth). We leverage the teacher forcing to accelerate training by taking the ground truth as the input for timestep $t+1$. 

After training the above predictor from real driving data, we can infer the probability distribution of each agent's future trajectories from time $t+1$ to $t+T$. Given the memory occupation and complexity, we only consider background agents in the Region Of Interest~(ROI) that may impact the ego agent. Note that the range of ROI is an important factor for the inference speed of the predictor. A large ROI means that LitSim needs to predict many agents.
\begin{table*}[htbp] 
    \centering
    \caption{Input features of control policy}
    \label{features}
    \setlength{\tabcolsep}{4mm}{
    \begin{tabular}{clcl}
        \toprule
        Group & Feature & Units & Description \\
        \midrule
    \textbf{ }    & Length   & \unit{m}  &  bounding box length of the ego agent \\
    \textbf{Core features }    & Height   & \unit{m}  &  bounding box height of the ego agent \\
    \textbf{ }  & Velocity  & \unit{m/s}   & longitudinal speed of the ego agent\\
    \textbf{ }    & Acceleration   & \unit{m/s^2}  & longitudinal acceleration, positive for acceleration, negative for deceleration  \\
        \midrule
    \textbf{ }  & Distance    & \unit{m}  & distance between the ego agent and the surrounding agent's front ends \\
    \textbf{Surrounding features }    & Velocity      & \unit{m/s}  & longitudinal speed of the surrounding agent \\  
    \textbf{ }    & Relative Angle      & \unit{rad}  & angle of the surrounding agents relative to the ego agent\\
    \textbf{ }    & Relative Heading Angle      & \unit{rad}  & heading angle of the surrounding agent relative to the ego agent \\ 
        \midrule
    \textbf{ }  & Marker Dist. (L)    & \unit{m}   &  center of the ego agent's lateral distance to left lane marking  \\  
    \textbf{ }    & Marker Dist. (R)  & \unit{m}   &  center of the ego agent's lateral distance to right lane marking \\
    \textbf{Road features }    & Distance Road (L)    & \unit{m}  & center of the ego agent's lateral distance to left road edge\\ 
    \textbf{ }    & Distance Road (R)    & \unit{m}   & center of the ego agent's lateral distance to right road edge \\
    \textbf{ }    & Lane Offset    & \unit{m}   & center of the ego agent's lateral centerline offset \\
    \textbf{ }     & Lane Curvature      & \unit{1/m}     & curvature of closest centerline point \\
    \textbf{ }    & Lane-Relative Heading    & \unit{rad}   &  center of the ego agent's heading angle relative to centerline \\
        \bottomrule 
    \end{tabular}
    }
\end{table*}
\subsection{Conflict Detection}
\label{Conflict Detection}
At each timestep, LitSim only reasons about the future trajectories of all agents in ROI. For background agents following replays, we obtain their future trajectories directly from the logs. LitSim checks for conflicts between the ego agent and background agents one by one, as well as conflicts among background agents themselves, based on their future trajectories. Note that a conflict does not always lead to a collision, as a collision only occurs when the simulated agent ignores the conflict. We follow~\cite{sun2022m2i} and define a conflict as the bounding boxes of two agents overlapping in both time horizon and spatial position. Once a potential conflict is detected, LitSim updates a new plan for the background agent. We provide an example of detecting a potential conflict, as shown in the first row of Fig.~\ref{pipeline}, where the yellow background agent will not yield to the red ego agent in the future.
\begin{figure}[t]
\centering
\includegraphics[width=\linewidth]{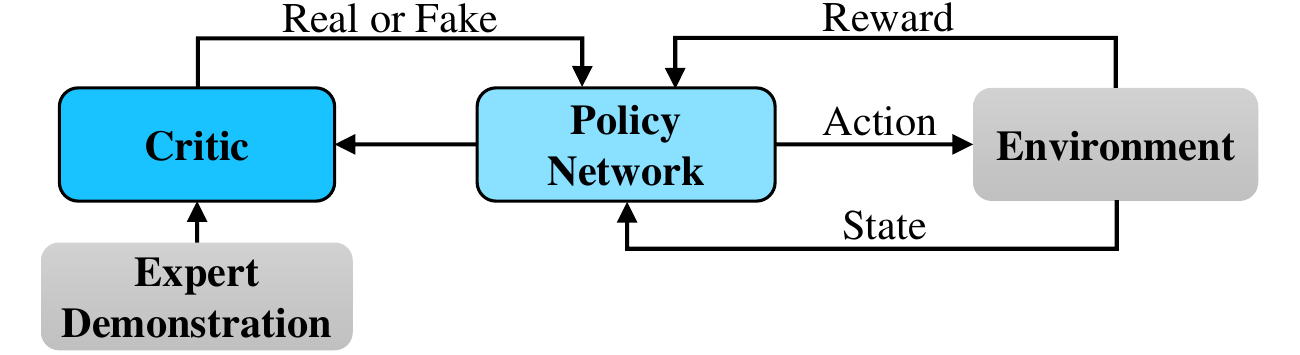}
\caption{{The pipeline of policy $C$}}
\label{policy}
\end{figure}
\subsection{Conflict-aware Control Policy}
\label{conflict-aware control policy}
We regard control policy $C$ as a sequential decision-making task. LitSim aims to train a policy $C$ from real-world data to achieve continuous dynamic control of agents. Our model is based on a critic and a policy network, and the pipeline is shown in Fig.~\ref{policy}. We build the policy network by rewarding it for "deceiving" a critic trained to discriminate between the policy output and expert data. The input of the critic comes from real expert data and the output of the policy network. The critic provides valuable reward signals to the policy network by discriminating the output of the policy network from expert data. The inputs to the policy network include the aforementioned critic's reward signal, externally provided reward, and observed states in the environment. Reward signals are obtained by the policy network imitating expert data, while the externally provided rewards are obtained from environmental interactions. Externally provided rewards specify the expert's prior knowledge, e.g., collisions with or proximity to other agents are undesirable. We use reward augmentation to combine imitation learning with reinforcement learning. Reward augmentation helps improve the learning agent's interaction and exploration of the state space. Additionally, we extract features from the environment related to the observed states as input to the policy network. The policy network maps the observed states into a high-dimensional vector and outputs the agent's decision-making behavior at the next timestep, such as yaw rate, longitudinal acceleration, and lateral acceleration. During training, we use Proximal Policy Optimization~(PPO)~\cite{schulman2017ppo} to update policy parameters until the critic is unable to discriminate between the expert data and the output of the policy network. 

We set a conflicting cross point as the agent's goal under the control policy $C$. Firstly, we obtain the connectivity relations of the roads where the agent is located and select a path to the target point using a graph search algorithm. We use a Bessel curve to roughly fit a smooth trajectory. Secondly, the policy $C$ achieves fine kinematic constraints for the agent based on the smooth trajectory, enabling the agent to reach the desired goal sooner or later than anticipated in the log. During the simulation, once a conflict is detected, policy $C$ takes over control of the agent. Then, the agent takes action continuously and reacts to new information, such as merge attempts and slowdowns. We follow the policy $L$ for background agents that will not conflict with other agents in the future. It means that background agents continue to replay the log until a conflict arises. Once the goal point is reached, the policy of the conflicting agent changes from $C$ to $L$. As shown in the second row of Fig.~\ref{pipeline}, the green background agent with policy $C$ brakes and yields to the red ego agent. After the conflict is resolved, the background agent continues to replay with policy $L$.

\subsection{Overall Simulation}
We evaluate the ego agent's performance by replaying logs and reasoning about future motion for all agents within the ROI at each timestep. The given future trajectories undergo scrutiny for potential conflicts between the ego agent and background agents. Then, the conflict will be resolved as described in Sec.~\ref{conflict-aware control policy}. On the other hand, if there are conflicts between background agents due to their deviation from the original trajectories, we iteratively execute Sec.~\ref{Conflict Detection} and Sec.~\ref{conflict-aware control policy} at each timestep until all conflicts are resolved.

\section{experiments}
\subsection{Experiment Setup}
\noindent\textbf{Predictor Training: }We train the predictor in Sec.~\ref{Joint Motion Prediction with Interaction} on NGSIM~\cite{NGSIM}, a real-world dataset containing highway and urban scenarios. We follow the experimental protocol of~\cite{social_lstm}, where the datasets are split into 70\% training, 10\% validation, and 20\% testing. We extract 8-second trajectories and use the first 3 seconds as history to predict 5 seconds into the future.

The first encoder encodes the map information into a rasterized representation. It consists of two 3 × 3 convolutional layers followed by a 2D max pooling layer. An additional 3 × 3 convolutional layer is applied, followed by an MLP with an output size of 64. The second and third encoders encode historical trajectories and current observed states into history vectors and interaction vectors, respectively. The second encoder consists of RNNs, each with 64 hidden layers. The third encoder consists of an MLP with an output size of 32 and an attention module with 8 parallel attention heads. In particular, both the RNNs encoder and decoder use bidirectional GRU, and the decoder has 128-dimensional hidden layers. We use ReLU as the activation function for all nonlinear functions.
\begin{table*}[htbp] 
    \centering
    \caption{Comparison against existing approaches}
    \label{expreiment result}
    \begin{tabular}{cccccccccc}
        \toprule
        {method} &ADE &ADE &ADE & ADE &ADE &Collision Rate & Reactivity rate&Progress &Relevant Ratio\\
        
        {} &@5\unit{s}(\unit{m})$\downarrow$   &@10\unit{s}(\unit{m})$\downarrow$   &@15\unit{s}(\unit{m})$\downarrow$   &@20\unit{s}(\unit{m})$\downarrow$  &@25\unit{s}(\unit{m})$\downarrow$   &(\%)$\downarrow$   &(\%)$\uparrow$  &(\unit{m})$\uparrow$   &(\%)$\downarrow$\\
        \midrule
    IDM~\cite{treiber2000congested}    &\textbf{0.25}   &\textbf{1.13}   &\textbf{2.56}   &4.67   &8.26   &8.68   &46.25   &192.25   &17.30   \\
    GAIL~\cite{ho2016generative}   &0.61   &2.78   &6.42   &12.36  &24.43   &6.76   &61.29 &\textbf{201.38}   &17.16  \\
    SimNet~\cite{bergamini2021simnet}   &0.52   &2.12   &5.74   &10.02  &18.68   &5.28   &80.35 &{197.50}   &14.20  \\
    \textbf{LitSim~(ours)}   &0.48   &1.37   &2.75   &\textbf{4.60}   &\textbf{7.73}   &\textbf{3.53}   &\textbf{91.36}   &195.03   &\textbf{2.40}   \\
        \bottomrule 
    \end{tabular}
\end{table*}
During the training, the predictor employs a variable batch size due to the fluctuating count of surrounding agents. We use an Adam optimizer to train the predictor. The initial learning rate is set at 0.001 and is decreased by a factor of 10 every 10 epochs.

\vspace{0.8em}\noindent\textbf{Policy $C$ Training: }We randomly select an ego agent from the NGSIM dataset and simulate its movements 300 steps ahead at a frequency of 10 Hz. The simulation will end early if the ego agent is involved in a collision or a traffic violation.

The policy network is composed of an MLP with an output size of 128. We divide the input features into three groups: core features of the ego agent, features of the surrounding agents, and road features. Their details are shown in Table~\ref{features}. 

During the training, we train the control policy $C$ in the gym~\cite{gym}, i.e., a reinforcement learning framework. We build the traffic simulation environment on SUMO. We leverage PPO to update model parameters during training time, whose learning rate is set at 0.03 and has a batch size of 32.

\vspace{0.8em}\noindent\textbf{Simulation Setup: }We use real traffic states from NGSIM to initialize the simulation. We split the dataset into 28-second long segments, each of which is with a frequency of 10 Hz. For each segment, the first 3 seconds serve as the initialization of the simulated state, and the last 25 seconds serve as the traffic scenario for the formal simulation.

To test the potential of LitSim for closed-loop evaluation, we use a state-of-the-art predictor called BAT~\cite{liao2024bat} to control the ego agent. It allows us to simulate realistic ego agent behaviors that closely follow the data distribution but deviate slightly from the logs. Consequently, these deviations require the simulator to modify the future trajectories of background agents to accommodate the altered ego agent's plan. Specifically, we randomly select an agent in the simulation and use BAT to control it.

\vspace{0.8em}\noindent\textbf{Baselines: } We choose three popular methods as our baselines, i.e., IDM, GAIL, and SimNet. IDM~\cite{treiber2000congested} is a heuristic model that generates simple following behavior via handcrafted rules. GAIL~\cite{ho2016generative} is an imitation learning model that learns human-like driving behavior by training a policy network with expert trajectory data. SimNet~\cite{bergamini2021simnet} uses a ResNet-50 model to simulate agents trajectories. In view of the randomness, we run five times for all approaches and choose the average values for comparison.

\subsection{Experimental Results}
We evaluate the performance of existing methods and LitSim in terms of the following metrics: 
\begin{itemize}
\item \textbf{Average Displacement Error~(ADE)}~\cite{caesar2021nuplan} stands for L2 distances between the trajectories of simulated agents and their ground truth trajectories up to the selected comparison horizon in the future. This metric is employed to evaluate the realism of simulation scenarios, with the objective being that an ideal closed-loop simulation should closely mirror real-world log data~\cite{bergamini2021simnet}\cite{surveydata}.

\item \textbf{Collision rate}~\cite{bergamini2021simnet} is defined as the ratio of colliding agents to all agents. If a collision lasts for multiple frames, we still count as a single collision. The \textbf{Reactivity rate}~\cite{bergamini2021simnet} is defined as the ratio of collision-free scenarios to the total number of scenarios. These two metrics measure the reactivity of a scenario~\cite{surveydata}.

\item \textbf{Relevant rate}~\cite{sun2022intersim} is defined as the ratio of taken-over agents to all agents. It measures the density of agents that need to be simulated in simulation.
\begin{figure}[htbp]
  \centering
  \includegraphics[scale=1.0]{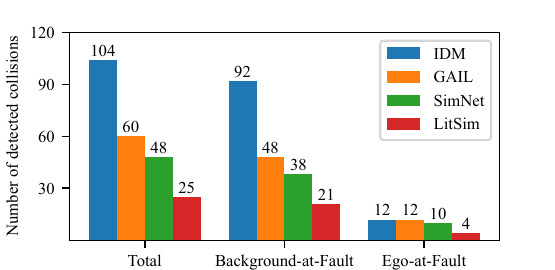}
  \caption{\textbf{The number of detected collisions between the ego agent and background agents.} In ego-at-fault scenarios where the ego agent is responsible for the accident, such as when it fails to yield at a stop sign and consequently collides with a background agent that has the right of way, the incident is classified as an "Ego-at-Fault" collision.}
  \label{histogram}
\end{figure}
\begin{figure*}[htbp]
\centering
\includegraphics[width=\textwidth]{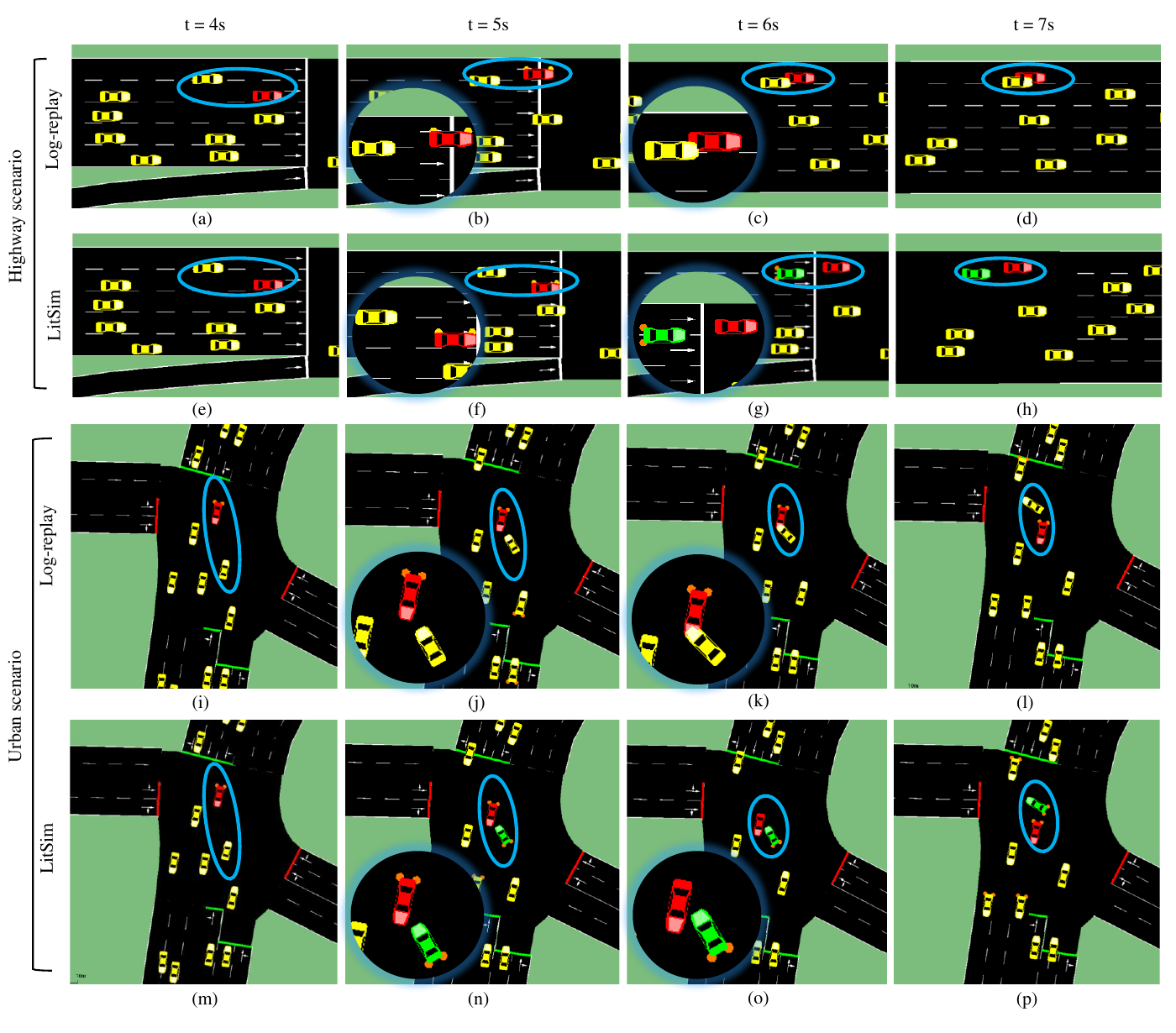}
  \caption{Two successful examples of LitSim. The \textcolor{red}{red} agent represents the ego agent, the  \textcolor{green}{green} background agent is under the control policy $C$, and the \textcolor{yellow}{yellow} background agents follow to log replay.}
\label{Sumo Simulation}
\end{figure*}
\item \textbf{Progress}~\cite{caesar2021nuplan} is defined as the average traveling distance of all agents. It measures agents' progress along the expert’s route ratio.
\end{itemize}
The quantitative results are shown in Table~\ref{expreiment result}. We summarize the results as follows.

(1) \textit{Realism:} In short-term simulations (up to 15 seconds), IDM excels in terms of ADE, slightly outperforming LitSim by a margin of 0.19. This advantage comes from IDM's straightforward behavior, which facilitates close following in dense traffic situations. In contrast, in long-term simulations (beyond 15 seconds), LitSim outperforms all baseline models in ADE by minimizing interventions in the expert trajectory.

(2) \textit{Reactivity:} LitSim outperforms IDM, GAIL, and SimNet in collision rate reduction, with decreases of 56\%, 48\%, and 33\%, respectively. Furthermore, in terms of reactivity rate, LitSim demonstrates improvements of 98\%, 49\%, and 14\% compared to IDM, GAIL, and SimNet, respectively. IDM excels in ADE but falls short in collision rate. A narrow following distance proves inadequate for handling unexpected situations, such as collisions that cannot be avoided through emergency braking due to kinematic constraints. SimNet focuses on generating individual future trajectories but lacks conflict checks. In contrast, LitSim predicts agents' future trajectories considering their interactions and spatial locations, using a conflict-aware policy for advanced conflict resolution. From a real-world perspective, our method provides human drivers with extended reaction times.

(3) \textit{Progress and relevant rate:} LitSim is better than other approaches in terms of relevant rate. ROI significantly impacts the relevant rate, and LitSim limits background agents within the ROI range. Although GAIL performs best in terms of progress, it causes the agents to deviate from the logs over long horizons. It is acceptable for Litsim to make less progress than GAIL because LitSim tends to yield to the agents.

In addition, Fig.~\ref{histogram} further analyzes the details of the resolved collisions. Fig.~\ref{histogram} shows that LitSim reduces the overall number of collisions by 76\%, 58\%, and 48\% compared to IDM, GAIL, and SimNet, respectively. Specifically, LitSim significantly reduces the number of Background-at-Fault collisions and Ego-at-Fault collisions compared to baselines. Although LitSim does not completely eliminate Background-at-Fault collisions, it greatly reduces their probability. After endowing background agents with intelligence, the overall driving environment becomes safer. This aligns with our common sense, indicating that collisions are less frequent among skilled drivers~\cite{rolison2018factors}.

\vspace{0.2em}\noindent\textbf{Take Away: }LitSim effectively balances realism and reactivity in long-term simulations, ensuring scene reactivity with minimal sacrifice to realism.
\begin{table*}[htbp]
    \vspace{5pt}
    \centering
    \caption{Ablation Study}
    \label{Ablation Study}
    \setlength{\tabcolsep}{4mm}{
    \begin{tabular}{ccccccccc}
        \toprule
        method & \textbf{Joint Prediction} & \textbf{Control Policy} & \textbf{ROI} & ADE & Collision Rate & Progress & Relevant Ratio\\
        {}     &  &  &(\unit{m})  &@25\unit{s}(\unit{m})$\downarrow$   &(\%)$\downarrow$  &(\unit{m})$\uparrow$  &(\%)$\downarrow$ \\  
        \midrule
    $\mathcal{M}_0$    & \ding{55} & \ding{55}      & 15  &46.88 &10.29 &\textbf{196.51} &2.79\\
    $\mathcal{M}_1$    & \checkmark   & \ding{55}   & 15  &9.62 &3.73 &194.79 &1.85\\
    $\mathcal{M}_2$    & \checkmark   & \checkmark  & 15  &8.63 &3.66 &195.87 &\textbf{1.74}\\
        \midrule
    $\mathcal{M}_*$   & \checkmark   & \checkmark  & 30   &\textbf{7.73} &\textbf{3.53} &195.03 &2.40\\
        \bottomrule 
    \end{tabular}
    }
\end{table*}
\subsection{Case Study}
We show two case studies in Fig.~\ref{Sumo Simulation}. As shown in the first two rows, we replay the log from US-101 highway. An unrealistic collision occurs~(Fig.~\ref{Sumo Simulation}(c)) due to the ego agent's lane-changing behavior being invisible to the background agent~(Fig.~\ref{Sumo Simulation}(b)). LitSim controls the background agent to slow down to resolve the conflict~(Fig.~\ref{Sumo Simulation}(g)). In the second two rows, we replay the log from Peachtree Street. In this unprotected left-turn scenario, the collision occurs~(Fig.~\ref{Sumo Simulation}(k)) even though the ego agent slows down~(Fig.~\ref{Sumo Simulation}(j)). LitSim takes over the background agent to yield~(Fig.~\ref{Sumo Simulation}(o)).

\subsection{Ablation Study}
We show the importance of each component of LitSim in Table~\ref{Ablation Study}. In the columns, a mark means whether a component of joint prediction or control policy is used in the simulation,~(i.e., a tick indicating Yes, a cross indicating No). We also investigated the impact of different ROI ranges in the simulation. Simulation~$(\mathcal{M}_0)$ performs poorly due to an inability to recognize and resolve the conflicts. Joint prediction~$(\mathcal{M}_1)$ is an important component in LitSim. It minimizes the need for intervention in original logs by accurately detecting conflicts, substantially narrowing the gap between simulation and reality. The conflict-aware control policy~$(\mathcal{M}_2)$ encourages interactions between agents to achieve fine-grained control of background agents. Despite controlling more background agents within a larger ROI~$\left(\mathcal{M}_2 \& \mathcal{M}_*\right)$, both ADE and collision rate experienced a reduction.

\subsection{Limitation}
While LitSim effectively mitigates the constraints associated with log replay, there remains a minor incidence of unrealistic collisions in the scenarios. Firstly, deep learning-based predictors encounter huge challenges in long-term prediction tasks, and they may suffer from misdetections or missed conflicts. Furthermore, the absence of history states for agents that abruptly enter the ROI presents an additional hurdle. Lastly, some collisions recorded in the collected data may arise from device accuracy.

\section{CONCLUSIONS}
In this work, we have proposed a novel method for generating interactive and realistic traffic simulations over long horizons. LitSim maximizes realism by minimizing the interventions in the raw log and generates consistent plans for all agents in the scene jointly. Experiments on NGSIM dataset confirmed that our method outperforms IDM, GAIL, and SimNet in terms of realism and reactivity. Future work will focus on expanding the diversity of scenarios and providing a control policy with alternative driving styles to improve the scalability of our method.


\bibliographystyle{IEEEtran}
\bibliography{ref-sample}
\addtolength{\textheight}{-12cm}   




\end{document}